\documentclass[conference]{IEEEtran}
\IEEEoverridecommandlockouts
\usepackage{cite}
\usepackage{amsmath,amssymb,amsfonts}
\usepackage{algorithmic}
\usepackage{graphicx}
\usepackage{textcomp}
\usepackage{xcolor}
\def\BibTeX{{\rm B\kern-.05em{\sc i\kern-.025em b}\kern-.08em
    T\kern-.1667em\lower.7ex\hbox{E}\kern-.125emX}}
\begin{document}

\title{Flexible electrical impedance tomography for tactile interfaces\\

\thanks{This work was supported in part by the European Research Council Starting Grant under Grant no.101165927 (Project SELECT).}
}

\author{\IEEEauthorblockN{Huazhi Dong}
\IEEEauthorblockA{\textit{SMART Group, Institute for} \\
\textit{Imaging, Data and Communications}\\
\textit{The University of Edinburgh}\\
Edinburgh, UK \\
huazhi.dong@ed.ac.uk}

\and
\IEEEauthorblockN{Sihao Teng}
\IEEEauthorblockA{\textit{SMART Group, Institute for} \\
\textit{Imaging, Data and Communications}\\
\textit{The University of Edinburgh}\\
Edinburgh, UK \\
sihao.teng@ed.ac.uk}

\and
\IEEEauthorblockN{Xiaopeng Wu}
\IEEEauthorblockA{\textit{SMART Group, Institute for} \\
\textit{Imaging, Data and Communications}\\
\textit{The University of Edinburgh}\\
Edinburgh, UK \\
xiaopeng.wu@ed.ac.uk}

\and
\IEEEauthorblockN{Xu Han}
\IEEEauthorblockA{\textit{SMART Group, Institute for} \\
\textit{Imaging, Data and Communications}\\
\textit{The University of Edinburgh}\\
Edinburgh, UK \\
s2143459@ed.ac.uk}

\and
\IEEEauthorblockN{Francesco Giorgio-Serchi}
\IEEEauthorblockA{\textit{Institute for Integrated Micro and Nano} \\
\textit{Systems}\\
\textit{The University of Edinburgh}\\
Edinburgh, UK \\
f.giorgio-serchi@ed.ac.uk}

\and
\IEEEauthorblockN{Yunjie Yang}
\IEEEauthorblockA{\textit{SMART Group, Institute for} \\
\textit{Imaging, Data and Communications}\\
\textit{The University of Edinburgh}\\
Edinburgh, UK \\
y.yang@ed.ac.uk}
}

\maketitle

\begin{abstract}
Flexible electrical impedance tomography (EIT) is an emerging technology for tactile sensing in human-machine interfaces (HMI). It offers a unique alternative to traditional array-based tactile sensors with its flexible, scalable, and cost-effective one-piece design. This paper proposes a lattice-patterned flexible EIT tactile sensor with a hydrogel-based conductive layer, designed for enhanced sensitivity while maintaining durability. We conducted simulation studies to explore the influence of lattice width and conductive layer thickness on sensor performance, establishing optimized sensor design parameters for enhanced functionality. Experimental evaluations demonstrate the sensor's capacity to detect diverse tactile patterns with a high accuracy. The practical utility of the sensor is demonstrated through its integration within an HMI setup to control a virtual game, showcasing its potential for dynamic, multi-functional tactile interactions in real-time applications. This study reinforces the potential of EIT-based flexible tactile sensors, establishing a foundation for future advancements in wearable, adaptable HMI technologies.
\end{abstract}


\section{Introduction}
Electronic skin (e-skin) is a flexible sensing technology that mimics human skin\cite{booth2018omniskins}. It enables robots to accurately perceive and interact with their environment, even in unstructured environments\cite{li2022multifunctional}. Recently, e-skin has gained significant attention in human-machine interfaces (HMI), health monitoring and wearable devices \cite{dai2024self, lai2023emerging, lu2023wearable}, and is regarded as one of the promising technologies for achieving embodied intelligence\cite{liu2024aligning}. Typically made of flexible and stretchable materials embedded with multiple sensors, e-skin is capable of sensing external pressure, temperature, humidity, and more. Among these, tactile sensors play a crucial role, which can detect contact positions and external forces. Currently, most tactile sensors employ an array structure comprising numerous independent sensing units\cite{wu2024bimodal},\cite{zhi2024hybrid}. While effective, these designs often involve complex structures and wiring, which not only increase the fabrication cost but also compromise sensor flexibility and scalability. Large-area sensing with array-based sensors also encounters limitations, such as poor scalability, high cost, low spatial resolution, and complex data communication requirements\cite{dahiya2019large, wang2023tactile}, which further limits their practicality in broader applications.

Electrical impedance tomography (EIT) has emerged as a promising alternative for achieving large-area tactile perception without array structures or invasive setups\cite{van2020large}. By placing sparse electrodes around the Region of Interest (ROI), EIT reconstructs the conductivity distribution within the ROI based on the measured voltages, allowing the estimation of contact positions and applied forces. Compared with traditional array sensors, EIT-based flexible tactile sensors adopt a “one-piece” structure without discrete internal components, making them low-cost and scalable. Furthermore, the use of ionic liquids\cite{chen2024correcting}, hydrogels\cite{park2022biomimetic} and elastic films\cite{wu2022new} as the conductivity layer enhances the stretchability, tactile sensitivity, and self-healing capabilities of the EIT-based tactile sensor. Recently, an ultra-thin EIT-based e-skin has demonstrated dynamic trajectory tracking of pressure points\cite{kim2024extremely}, verifying EIT's high spatiotemporal resolution and continuous sensing capability.

However, conventional EIT sensors with uniformly distributed conductive paths exhibit relatively small conductivity changes under local pressure, which influences the sensor’s sensitivity. To address this issue, recent research has explored porous\cite{chen2022large} and multi-layer structures\cite{park2021deep} to improve sensitivity, though those often involve complex fabrication. In contrast, lattice-patterned conductive paths offer excellent repeatability and improved spatial resolution, while maintaining great linearity over a wide pressure range\cite{yin2021facile}. In addition, the lattice channels can carry a higher current density, leading to more significant and faster conductivity changes under deformation\cite{jamshidi2023eit}. 

Here, we introduce a novel flexible tactile sensor with a 3D lattice-pattened structure, achieving high resolution, sensitivity and durability. The key contributions of this work are as follows:
\begin{itemize}
    \item The modelling and design of an EIT-based tactile sensor with a lattice pattern, which is fabricated with hydrogel and silicone, and the experimental validation of its tactile reconstruction ability. 
    \item The implementation and testing of the tactile sensor's enhanced sensitivity and response time in a practical HMI application.
\end{itemize}

The remainder of this paper is organized as follows: Section II discusses the principles of EIT-based tactile sensing, Section III describes the fabrication of the lattice-shaped tactile sensor, Section IV presents tactile reconstruction results in real-world experiments and HMI applications, and Section V concludes the paper with discussions on future research directions.

\section{Principle of EIT-based tactile sensing based on EIT}
\begin{figure}[t]
\centering
\includegraphics[width=\columnwidth]{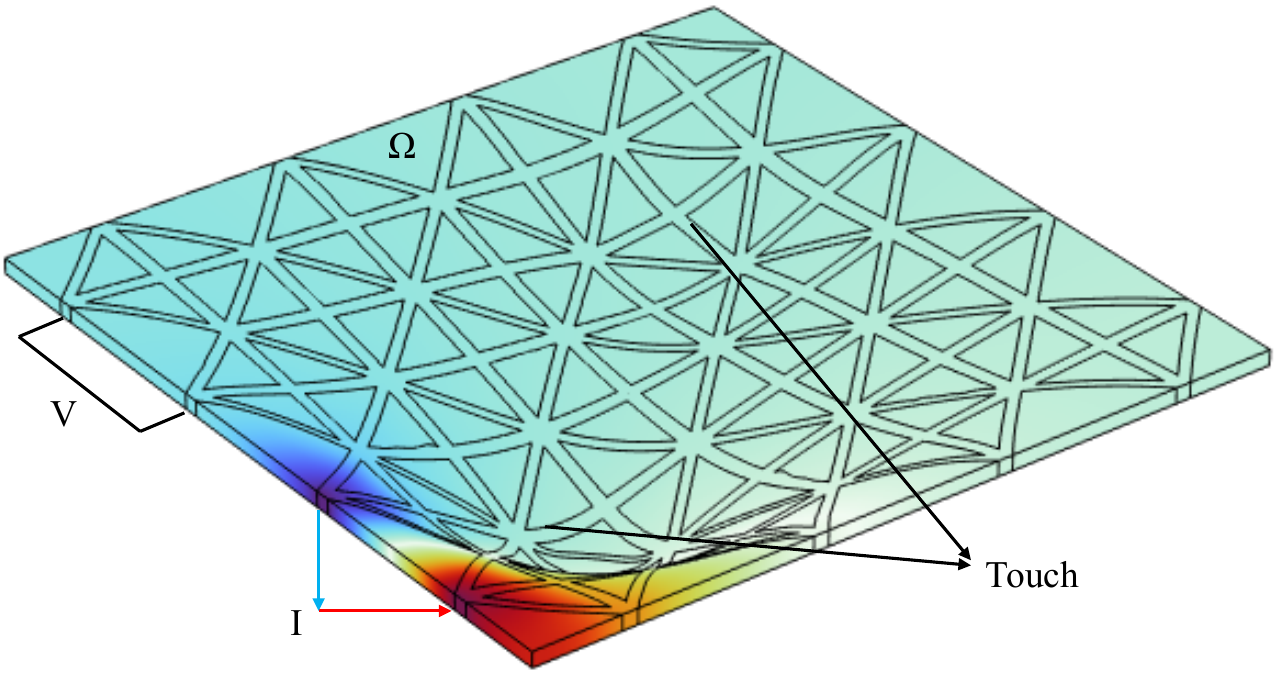}
\caption{Schematic illustration of EIT-based tactile sensing.}
\label{fig-eittheory}
\end{figure}
The principle of EIT-based tactile sensing is illustrated in Fig. \ref{fig-eittheory}. When pressure is applied locally to a sample substrate, it induces localized changes in conductivity in the touched areas, which in turn affect the electric potentials at the boundary electrodes. Here, the goal of EIT-reconstruction is to estimate the continuous conductivity distribution $ \boldsymbol{\sigma}$ in the sensing region $ \boldsymbol{\Omega} $ based on the voltage measurements $ \boldsymbol{V} $. EIT involves forward and inverse problems. The forward problem predicts surface potentials for a known conductivity and applied current. This relationship is governed by:
\begin{equation}
    \nabla \cdot \left [{ \sigma \left ({x,y }\right)\nabla u\left ({x,y }\right) }\right]=0,\quad \left ({x,y }\right)\in \boldsymbol{\Omega} \tag{1}
    \label{eq1}
\end{equation}
where $\sigma \left ({x,y }\right)$ represents the conductivity at $\left ({x,y }\right)$, and $u \left ({x,y }\right)$ is the potential distribution. This can be further linearized as follows:
\begin{equation}
    \Delta \boldsymbol{V}=\boldsymbol{J}\Delta\boldsymbol{\sigma}\tag{2}
    \label{eq2}
\end{equation}
where $\Delta V $ denotes the voltage change, and $ \Delta \boldsymbol{\sigma} $ is the conductivity change. The EIT inverse problem involves estimating $ \Delta \boldsymbol{\sigma} $ from $\Delta V $, which is typically formulated as a regularized optimization problem:
\begin{equation}
    \displaystyle \arg\min_{\Delta\boldsymbol{\sigma}}\frac{1}{2}\Vert \boldsymbol{J}\Delta\boldsymbol{\sigma}-\Delta \boldsymbol{V}\Vert_{2}^{2}+\lambda R(\Delta\boldsymbol{\sigma})\tag{3}
    \label{eq3}
\end{equation}
where $R$ is the regularization term incorporating prior knowledge, and $\lambda>0$ is the regularization factor that balances data fidelity and regularization. The conductivity reconstruction forms the basis for detecting tactile interactions through EIT-based tactile sensors.

\section{Lattice-patterned Flexible Tactile Sensor}
\subsection{Simulation}
\begin{figure*}[t]
\centering
\includegraphics[width=\textwidth]{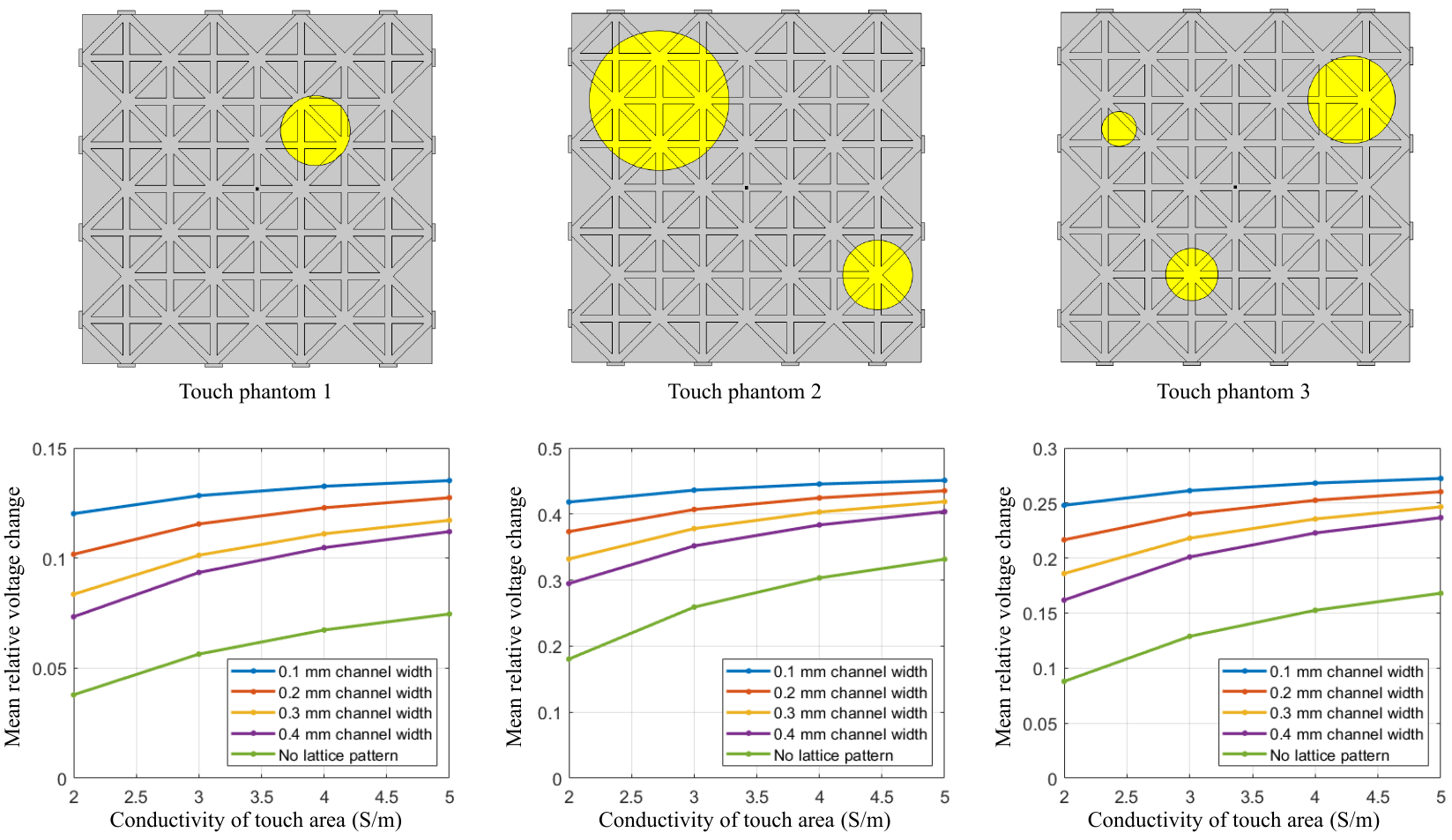}
\caption{Effect of channel widths on EIT measurements using 2D simulation. Yellow represents touch areas. The conductivity of the untouched area is 1 S/m.}
\label{fig-2dsim}
\end{figure*}
\begin{figure*}[!th]
\centering
\includegraphics[width=\textwidth]{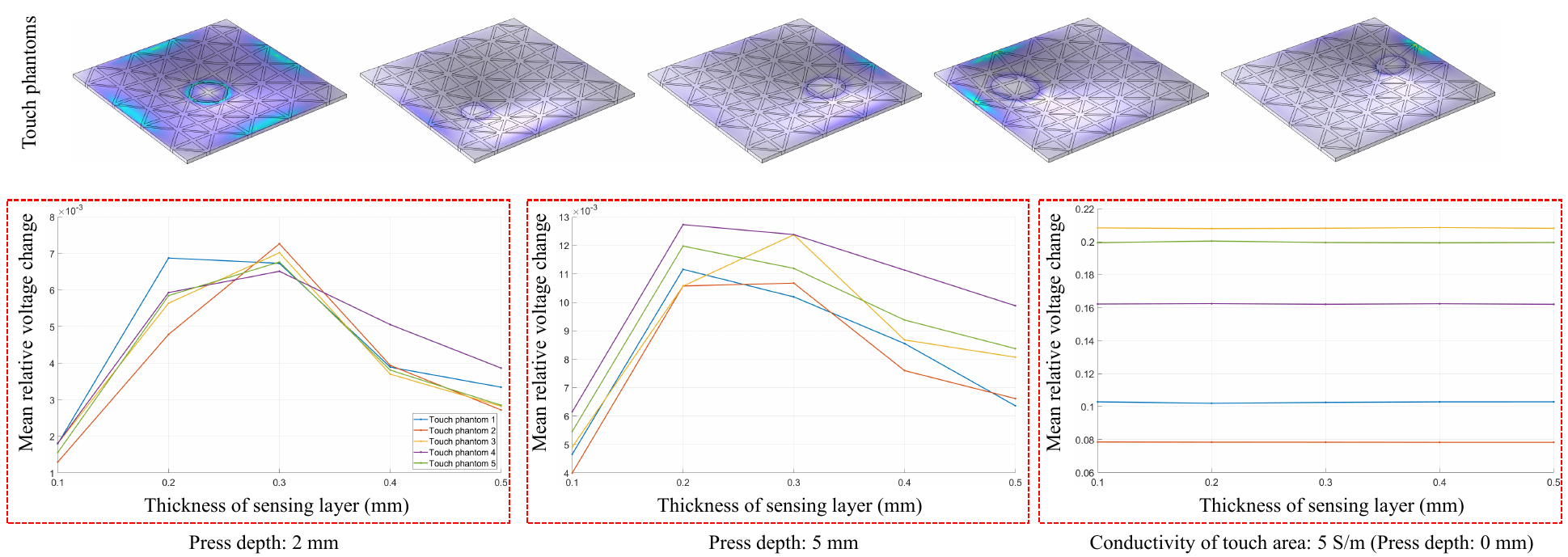}
\caption{Effect of sensing layer thicknesses on EIT measurements using 3D simulation. Yellow represents touch areas. The conductivity of untouched area is 1 S/m.}
\label{fig-3dsim}
\end{figure*}
To evaluate and optimize the sensor’s sensitivity, we simulated five different lattice channel widths. In these simulations, changes in conductivity indicated tactile interactions. As shown in Fig. \ref{fig-2dsim}, we tested three touch phantoms, i.e., single, double and triple touch points, at four distinct conductivity levels. Using the \textit{adjacent-driven adjacent-measurement} protocol \cite{brown1987sheffield}, we obtained 104 EIT measurements. The results indicate that the average relative change in measured voltages increases with the conductivity of the tactile region. Notably, the lattice structure consistently yields a higher voltage variation than the non-lattice structure, confirming its enhanced sensitivity. Moreover, reducing the channel width within the lattice structure further amplifies its sensitivity.

Based on these findings, we conducted a coupling field simulation \cite{hu2023coupling} to model more realistic tactile interactions and assess the influence of conductive layer thickness on sensor sensitivity. This setup tested five different thicknesses for the conductive layer and five distinct touch locations. As shown in Fig. \ref{fig-3dsim}, we applied three different levels and types of pressure to the touch area: a 2 mm press, a 5 mm press, and a local conductivity increase to 5 S/m. The results demonstrate that, under identical touch conditions and positions, increasing the conductive layer thickness has minimal impact on sensor sensitivity.

Our previous study \cite{dong2024tactile} identified hydrogel as a suitable material for the conductive layer due to its biocompatibility and flexibility. Based on the simulation results and practical considerations, the final sensor design incorporates a hydrogel lattice pattern with a conductive layer thickness of 2 mm and a channel width of 2 mm, optimized for enhanced tactile responsiveness.

\subsection{Sensor Fabrication}
\begin{figure*}[!ht]
\centering
\includegraphics[width=\textwidth]{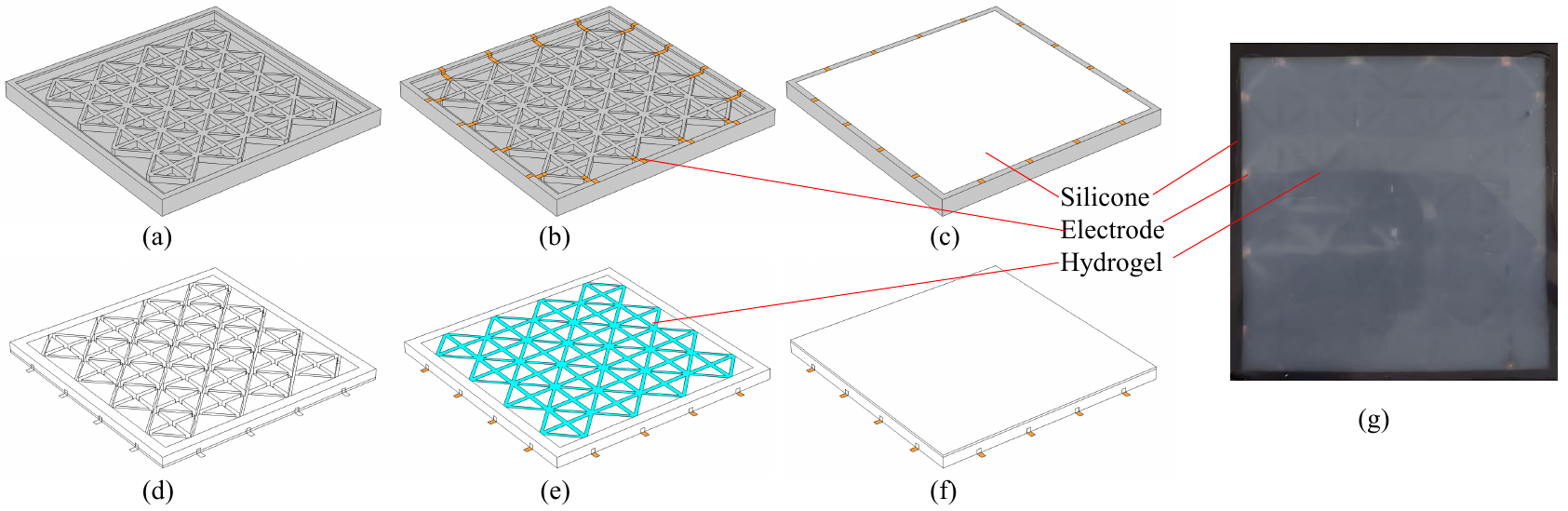}
\caption{Fabrication process of the Lattice-patterned tactile sensor. (a) 3D-printed mould. (b) Deploy electrodes on the mould. (c) Eco-flex was poured into the mould. (d) Cure at room temperature for 4 hours and release the mould. (e) The pre-gel solution of hydrogel was poured onto the cured silicone and polymerized by exposing it to UV light (365 nm) for 2 hours. (f) Silicone was poured over the entire sensor to seal and prevent environmental interference. (g) The fabricated Lattice-patterned tactile sensor. }
\label{fig-sensor}
\end{figure*}
\begin{figure*}[!ht]
\centering
\includegraphics[width=\textwidth]{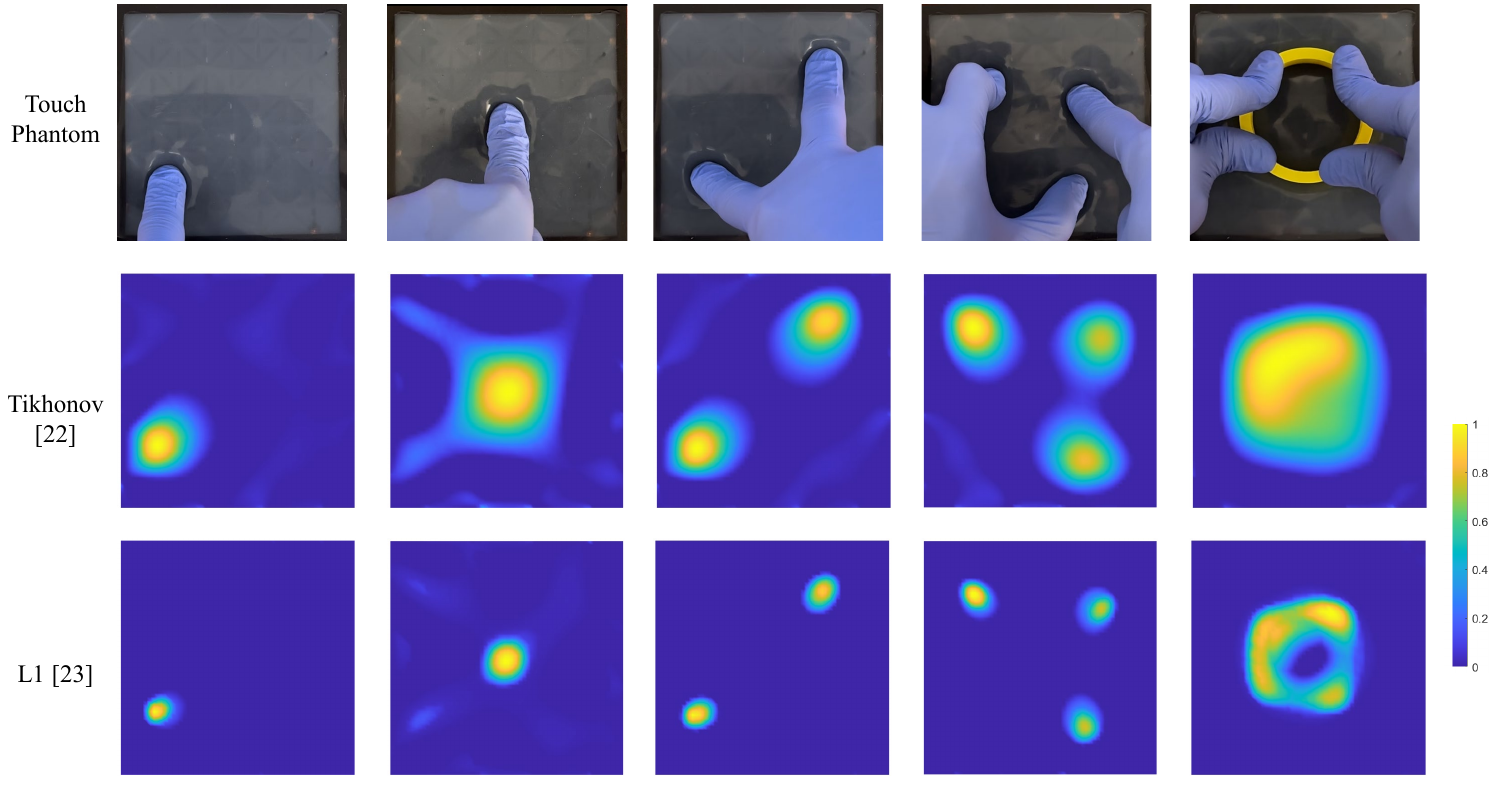}
\caption{Tactile reconstruction using experiment data. To ensure fair comparison, images were normalized to have a maximum value of 1.}
\label{fig-recon}
\end{figure*}
Fig. \ref{fig-sensor} shows the fabrication process of the EIT-based lattice-patterned tactile sensor. The sensor comprises a substrate layer, a sensing layer, a sealing layer, and 16 evenly distributed boundary electrodes. The sensor's overall dimensions are 110 $\times$ 110 $\times$ 4 mm$^{3}$, with the conductive layer positioned within the sensor measuring 100 $\times$ 100 $\times$ 2 mm$^{3}$. Each electrode forms a 3 $\times$ 3 mm$^{2}$ contact area with the conductive layer to ensure effective signal transmission. The step-by-step fabrication process is as follows:
\begin{enumerate}
\item A 3D-printed mould is prepared to define the sensor structure.
\item Copper electrodes are carefully placed within the mould.
\item A pre-gel mixture of silicone is poured into the mould. This mixture is prepared by combining parts A and B of Ecoflex 00-30 (Smooth-On Inc.) in a 1:1 ratio. A black pigment is added to the silicone for enhanced visual contrast.
\item After allowing the silicone to cure for 4 hours at room temperature, the substrate layer is released from the mould.
\item The hydrogel precursor is then poured into the designated sensing area and cured under UV light (365 nm) for two hours, following the methodology outlined in our previous study \cite{dong2024tactile}.
\item A final layer of silicone pre-gel is applied over the entire sensor to form a sealing layer, effectively protecting it from environmental interference.
\end{enumerate}

This method yields a robust and flexible tactile sensor with high sensitivity and durability, making it well-suited for tactile sensing and various human-machine interaction applications.

\section{Results and Discussion}

\subsection{Tactile Reconstruction}
In the experiments, a current frequency of 10 kHz was employed, following the adjacent-driven adjacent-measurement protocol. Five distinct tactile phantoms were evaluated, including single-point, double-point, triple-point, and annular touch configurations. To reconstruct the tactile images, we employed Tikhonov \cite{lionheart2004eit} and L1 regularization \cite{tehrani2012l1} algorithms. Both methods were optimized through extensive parameter tuning to achieve the best possible performance. For the Tikhonov regularization, a regularization factor of 0.01 was selected. Similarly, the L1 regularization utilized a regularization factor of 0.01, with the number of iterations set to 200. To focus solely on tactile presses in one direction, values of the tactile reconstruction below zero were disregarded in all results.

As illustrated in Fig. \ref{fig-recon}, the reconstructed images accurately capture the positions of all tactile inputs. Notably, the annular touch configuration is precisely reconstructed in its shape using the L1 regularization algorithm. These results demonstrate the efficacy of the designed sensor in capturing various tactile interactions, validating its suitability for various tactile sensing applications.

\subsection{HMI Application}
\begin{figure}[t]
\centering
\includegraphics[width=\columnwidth]{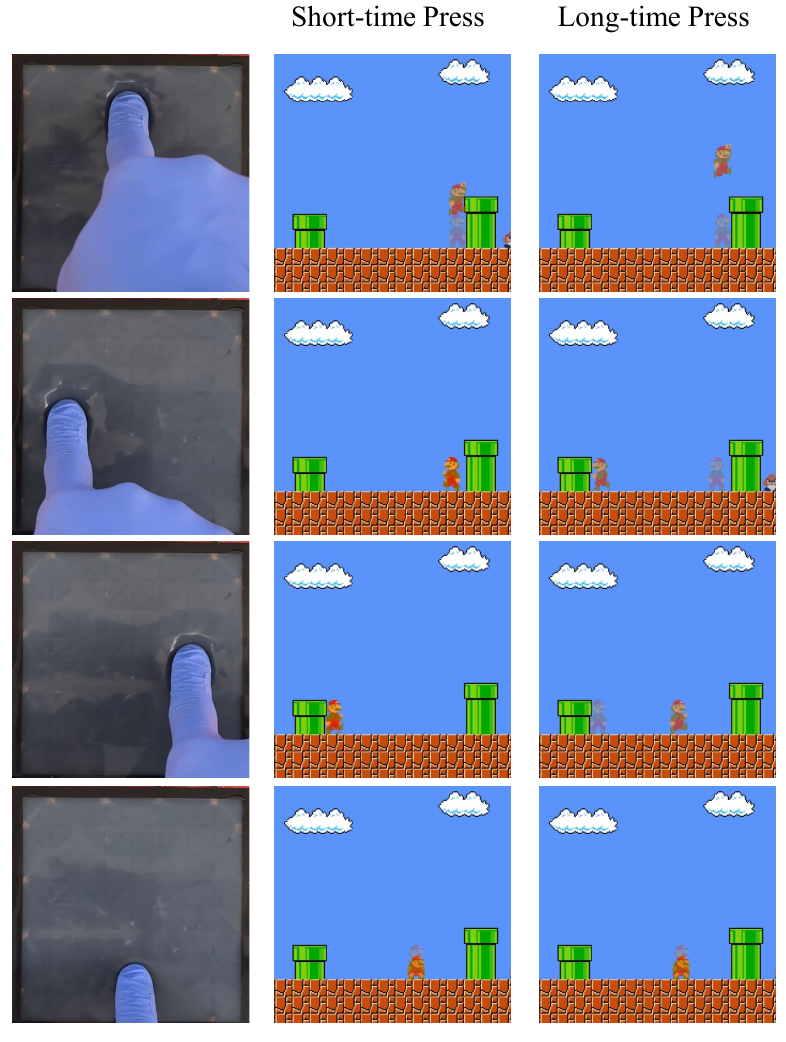}
\caption{HMI application: Super Mario Bros Game.}
\label{fig-HMI1}
\end{figure}
The designed tactile sensor was further evaluated as an HMI interface in a practical application to control a virtual character in the Super Mario Bros Game, as shown in Fig. \ref{fig-HMI1}. The left column shows the physical tactile contact on the sensor, and the right column describes the actions generated in the virtual environment, illustrating the character's response to each tactile input. When the user presses the sensor in different locations, the corresponding action is performed in the game, such as advancing or jumping. When the user presses the sensor for a different length of time, it will also be recognized as a different motion amplitude, such as low-altitude jumping and high-altitude jumping. The supplementary video shows the real-time control for the Super Mario Bros Game. This demonstrates the feasibility of using our tactile sensors for real-time, intuitive control in virtual environments, underscoring their ability to capture and transform subtle tactile inputs for multi-functional HMI applications.

\section*{Conclusion}
This study presents a flexible, lattice-patterned EIT-based tactile sensor as a promising solution for tactile sensing applications. By leveraging a hydrogel-based conductive layer, the sensor achieves high spatial resolution, sensitivity and robustness, while maintaining biocompatibility and adaptability, which are key features for human-machine interface and wearable technology applications. Our future research will focus on refining the sensor’s design for scalability, optimizing it for larger-area applications, and exploring multi-modal sensory integration to broaden its functional scope in diverse and challenging environments.

\bibliographystyle{IEEEtran}
\bibliography{References}

\end{document}